%% file: main.tex
\documentclass[10pt,twocolumn,letterpaper]{article}

\usepackage[]{cvpr}      
\input{preamble}
\definecolor{cvprblue}{rgb}{0.21,0.49,0.74}
\usepackage[pagebackref,breaklinks,colorlinks,allcolors=cvprblue]{hyperref}

\def\paperID{*****} 
\def\confName{CVPR}
\def\confYear{2026}

\title{A Light Weight Multi-Features-View Convolution Neural Network For Plant Disease Identification}

\author{Muhammad Kaleem Ullah Khan\\
COMSATS University Islamabad, Pakistan\\
}

\begin{document}
\maketitle
\input{sec/0_abstract}    
\input{sec/1_intro}
{
    \small
    \bibliographystyle{ieeenat_fullname}
    \bibliography{main}
}


\end{document}

%% file: sec/0_abstract.tex
\begin{abstract}
Agriculture is a key sector of the economies of developing countries. It serves as a primary source of income and employment for rural populations. However, each year, a large portion of crops is wasted because of pests and diseases. Well-timed prediction of plant diseases is crucial to sustainable, high-quality agricultural production. Detection of plant diseases through conventional methods is both labour-intensive and time-consuming. Researchers have developed image classification based automated techniques for this purpose. Most accurate methods are based on deep convolutional neural networks, which are computationally intensive, with many layers and millions of trainable parameters. 
In resource-constrained settings, especially in rural areas, it is difficult to deploy deep convolutional neural network models for efficient plant disease identification. To address these issues, an efficient and light-weight Multi-View Convolutional Neural Network is proposed. 
These additional features aid the proposed model to identify the plant diseases accurately and efficiently with less number of parameters. The proposed model is tested on a benchmark Plantvillage dataset and achieves an improvement of $ 2.9\%$ in classification accuracy compared to the baseline convolutional neural network model, which was trained only on Red, Green, and Blue (RGB) plant images. Compared with state-of-the-art deep convolutional neural network models, the proposed model is less computationally expensive and achieves comparable accuracy for plant disease identification on the PlantVillage dataset.
\end{abstract}

%% file: sec/1_intro.tex
\section{Introduction}
\label{intro}

Agriculture plays a vital role in the world’s economy, contributing 6.4\% to global economic production \cite{abc}. Agriculture feeds billions of people worldwide, and some economies depend entirely on it. Farmers do backbreaking work planting and harvesting crops; however, their efficiency is affected by problems such as weed growth, pests, and plant diseases. These problems mostly go unnoticed and cause significant loss to the farmer.
\par
Most conventional farming methods involve farmers using excessive amounts of toxic chemicals to manage pests and plant diseases. These extreme uses of pesticides threaten farmers' and consumers' health, kill natural enemies of the plants, and disturb the overall ecosystem. Moreover, the use of insecticides also results in an upsurge in the price of the product. Therefore, early prediction of crop diseases and pests is vital for responding to the increasingly severe situation caused by major pests and diseases. At present, the commonly used prediction methods rely on data collected through field visits conducted by plant protection staff. The data is then sampled and analyzed. The problem with this process is that data collection and analysis involve a time lag, which directly affects the accuracy of prediction systems. According to a study, ill-timed pest and disease predictions are destroying more than 40\% of the crop each year, even with the use of 3 million tons of pesticides and non-chemical control measures, and improved varieties \cite{pimentel2009pest}.

Over the last decade, smart farming has gained popularity to address issues with conventional farming methods. Smart farming is essential to increase production and save the fields from pests and diseases. Early identification of pests and diseases is the cornerstone of sustainable and secure agricultural production. Smart farming uses machine vision and image processing to monitor plants. Liu \textit{et al.} \cite{liu2017review} reviewed different methods used for identifying invertebrates on crops such as slugs, locusts, butterflies and snails. Artificial intelligence, particularly neural networks, has become popular in image processing for detecting ripe fruits.  The accuracy between 60\% and 100\% in most cases depends on the nature of the fruit and other situations \cite{kurtulmus2014immature,cubero2016automated,chaivivatrakul2014texture}. Histogram of Oriented Gradients (HOG) is one of the best handcrafted feature extraction methods \cite{dalal2005histograms}, along with the Scale-Invariant Feature Transform (SIFT) \cite{lowe2004distinctive}. Various methods have been proposed for plant disease identification and segmentation, including decision trees, neural networks, and Support Vector Machine (SVM) based techniques \cite{petrellis2018review}.

Deep convolutional neural networks (DCNNs) revolutionized computer vision, achieving landmark results when they won the ImageNet competition in 2012 by halving the error rate \cite{krizhevsky2012imagenet}. In the coming years, the performance of DCNNs continued to improve, and they dominated ImageNet competitions. DCNNs have achieved significant results on many benchmark datasets comprising millions of images \cite{he2016deep,howard2017mobilenets}. 
Too \textit{et al.} \cite{too2019comparative} used several deep learning models, including VGG, ResNet, DenseNets, and Inception-V4, for timely and accurate plant disease prediction. 
Girshick \textit{et al.} \cite{girshick2015fast} proposed the Fast R-CNN model, which is a real-time deep learning technique. It improves object detection accuracy and reduces training and testing time. Fast R-CNN relies on proposal regions; when they are excluded, the network's performance decreases \cite{bousetouane2016fast}.
Esteva \textit{et al.} \cite{esteva2017dermatologist} proposed an end-to-end DCNN model for skin lesion classification and tested its effectiveness against 21 dermatologists on two tasks.  The proposed model demonstrated performance comparable to that of experts, and some researchers suggested that this work is deployable on mobile devices in real-world settings. With more evaluation \cite{ramcharan2017deep, johannes2017automatic}, this model also shows good classification results on plant diseases.

One problem with DCNNs is that they require a large amount of annotated data to be effective. However, in domains such as agricultural imaging, the availability of large labelled datasets is rare. Also, State-Of-The-Art (SOTA) CNNs are computationally expensive, requiring long training and inference times. 
For an efficient plant disease identification system, a compact, fast-matching system is required.
The proposed research presents a method to reduce the computational cost of DCNNs for plant disease identification, thereby addressing the above-mentioned issues. In the proposed methodology, we compute additional image information, such as image gradients, and pass it to the network along with the input image. We call this method multi-view CNN (MV-CNN). This extra information will help the CNN model to converge faster on large datasets without deeper layers. The reduced number of layers results in faster training and inference time, while the additional information helps retain network accuracy. The proposed changes will make the overall model lightweight without losing performance. The key contributions of the proposed work are given as follows:
\begin{enumerate}
    \item For the plant disease classification task, an MV-CNN model is proposed, and SOTA classification methods are compared with the proposed model on publicly available datasets.
    \item This method uses multiple views of an input image, such as image gradients, to speed up the network's learning.
    \item The proposed MV-CNN uses only seven convolution layers that have less complexity than SOTA classification models without significantly reducing accuracy.
    \item The performance of the proposed MV-CNN is evaluated against various SOTA classification models based on training accuracy, validation accuracy, number of parameters and processing time.
    \item Performance of this proposed model is tested by calculating the AUC-ROC curves against each class.
    \item Heat maps of every class are also calculated using the GradCam method to analyze the network's performance.
\end{enumerate}
The paper is organized as follows: background information
and related work on plant disease identification are discussed in Section II. Section III describes the proposed method, Section IV presents the results, and Section V concludes this paper.

\section{Related Works}
\label{sec:1}

The identification of plant diseases and pests has become increasingly important over the years. This research's main motivation is to develop an efficient plant disease identification system that farmers can use to predict plant diseases promptly. However, some elements are important to consider for the system's success, such as its cost-effectiveness, ease of use, sensitivity and accuracy \cite{fuentes2017robust}. Over the past decade, many studies have proposed several non-destructive techniques to address these issues. Meron \textit{et al.} \cite{meroni2008assessing} used hyperspectral proximal sensing techniques to assess environmental stresses in plants. Optical methods are practical tools for studying plant health \cite{liew2008signature}. These methods include thermal and fluorescent imaging techniques that have been introduced to estimate, among other things, increased plant stress caused by increased gases, radiation, water conditions, and insects. Another vital area is the study of plant protection against pathogens \cite{Gardener2014}.
DCNNs have recently been shown to be an effective technique for accurate plant disease prediction \cite{too2019comparative}. The introduction of DCNN methods enabled researchers to design accurate classifiers that do not require handcrafted feature extraction in an end-to-end manner. Farmers can effectively use these systems with minimal background knowledge or training. 

\subsection{DEEP CONVOLUTIONAL NEURAL NETWORK}
DCNN methods automatically learn patterns and extract features from input images for pixel-level localization and classification. A DCNN is a combination of layers and operations stacked on top of each other. The most commonly used DCNN architectures are composed of the following layers.

\begin{enumerate}

\item{\textit{Convolution layer}:
Convolution is the main building block of a DCNN model. In the convolutional layer, the input image is convolved with a filter to produce a feature map that captures the important features of the input. To introduce nonlinearity in the layer's output, different activation functions, such as the Rectified Linear Unit (ReLU), are applied to the feature maps.}
\item{\textit{Pooling layer}: 
The pooling layer takes the feature maps from the convolution layer and combines the semantically related features into a single feature. The features are usually combined by taking the average or maximum of a patch of feature values. This layer also reduces the computational complexity of the DCNN.}
\item{\textit{Fully connected layer}:
A fully connected layer is used to classify the feature vector extracted from the convolution layers. In this layer, each unit is fully connected with the units of the previous layer. Fully connected layers are usually added after stacking convolution and pooling layers multiple times.}
\item{\textit{Softmax layer}:
SoftMax is an activation function that converts feature values into output probabilities for different classes.}

\end{enumerate}

DCNNs are trained to minimize misclassifications by updating their model parameters. Various optimization algorithms, such as stochastic gradient descent, are used to update model parameters using the standard back-propagation algorithm.

Parwara \textit{et al.}\cite{pawara2017comparing} compared the performance of local feature descriptors, bag of visual words, and CNN-based methods for recognizing plants from multiple datasets. The authors found that the CNN-based approach outperformed the other techniques. Cugu \textit{et al.} \cite{cugu1701novel} extracted features using both the handcrafted methods and CNN to achieve highly accurate results for classifying 57 different species of trees using leaves. The authors trained an SVM for classification. The author proposed a deep learning-based architecture to identify banana leaf disease across different weather conditions and under various image characteristics, including complex backgrounds, image size, image quality, image pose or orientation, and illumination. LeNet architecture is also used in this paper to classify the datasets \cite{amara2017deep}.

Mohanty \textit{et al.} \cite{mohanty2016using} used two CNN-based networks, namely GoogleNet \cite{szegedy2015going}, and AlexNet \cite{krizhevskyimagenet}, for the identification of 26 plant diseases for 14 different crops using the Plantvillage dataset \cite{hughes2015open}. Similarly, Wang \textit{et al.} \cite{wang2017automatic} also used the same dataset to train a VGG-16 model and achieved 90.4\% test accuracy \cite{wang2017automatic}.
Fuentes \textit{et al.}\cite{fuentes2017robust} used three different types of CNN architectures to detect plant diseases. These methods include a single-shot multibox detector (SSD), a region-based fully convolutional network (R-FCN), and Faster R-CNN. The authors combined each architecture with Residual Networks (ResNets) and VGG-Net and applied local and global augmentations to increase the accuracy of tomato plant disease identification.
Arsenovic \textit{et al.} \cite{arsenovic2019solving} used different pre-trained models and performed experiments on the standard Plantvillage dataset\cite{Dataset}. The authors also generated 79265 additional plant images using Style-GAN and achieved 93.5\% accuracy.

Lee \textit{et al.} \cite{lee2020new} performed experiments on the Plantvillage dataset using four different architectures, namely GoogLeNetBN, InceptionV3, VGG16, and GoogLeNet, with 3 different configuration settings. In the first configuration, the authors performed classification using pre-trained weights from Imagenet \cite{imagenet_cvpr09}. Secondly, the authors performed classification by pre-training the model on the PlantCLEF2015 dataset \cite{LifeCLEF2015Planttask} and then fine-tuning it on the Plantvillage dataset. Finally, in the third configuration, the authors trained the model from scratch. The authors concluded that the pre-trained models performed better than training from scratch.

Zhang \textit{et al.} \cite{zhang2019three} introduced a simple CNN-based model to identify plant diseases in the Plantvillage dataset\cite{Dataset}. The authors passed each channel of the RGB image to a different CNN. After feature extraction, the outputs of the three CNNs were concatenated and fed into a single fully connected network for disease identification, achieving good accuracy.
Geet \textit{et al.} \cite{geetharamani2019identification} proposed a nine-layer CNN model and performed experiments on the Plantvillage dataset \cite{Dataset} with six different types of data augmentations. The authors achieved 96.46\% accuracy, comparable to SOTA models.

Although the methods mentioned above are effective when data availability is not an issue, handling small datasets or imbalanced classes across multiple applications remains a challenge \cite{fuentes2018high}. Furthermore, these models require substantial training time due to their highly complex, deep nature.
Therefore, it is important to develop strategies specifically designed for scenarios with limited or imbalanced data, and to design a method that can perform complex classification tasks at lower computational cost.

This study proposes a novel MV-CNN model that leverages additional information, including image gradients and the input image, for plant disease identification. With additional information, the proposed model can converge faster on large datasets without deeper layers, resulting in faster training and inference times. The accuracy of the proposed model is comparable to SOTA models. The following section describes the proposed MV-CNN and the overall framework for an efficient plant disease identification system.

\section{Framework}

The detailed diagram of the proposed MV-CNN model is shown in Figure 1. The MV-CNN architecture is divided into four subparts. Firstly, we calculate the image gradients in horizontal and vertical directions as well as the gradient magnitude of the input colour (RGB) image. At the second step, these pre-calculated features are concatenated with the input and given to the Multi-Feature Extraction Module (MFEM) for feature extraction using several convolutional layers. After that, the MFEM output is passed to the Feature Selection Module (FSM), which consists of fully connected layers for selecting relevant features. Finally, the classification is performed using the Prediction Module (PM).  

\begin{figure*}[!h]
	\centering
	\includegraphics[width=1\linewidth]{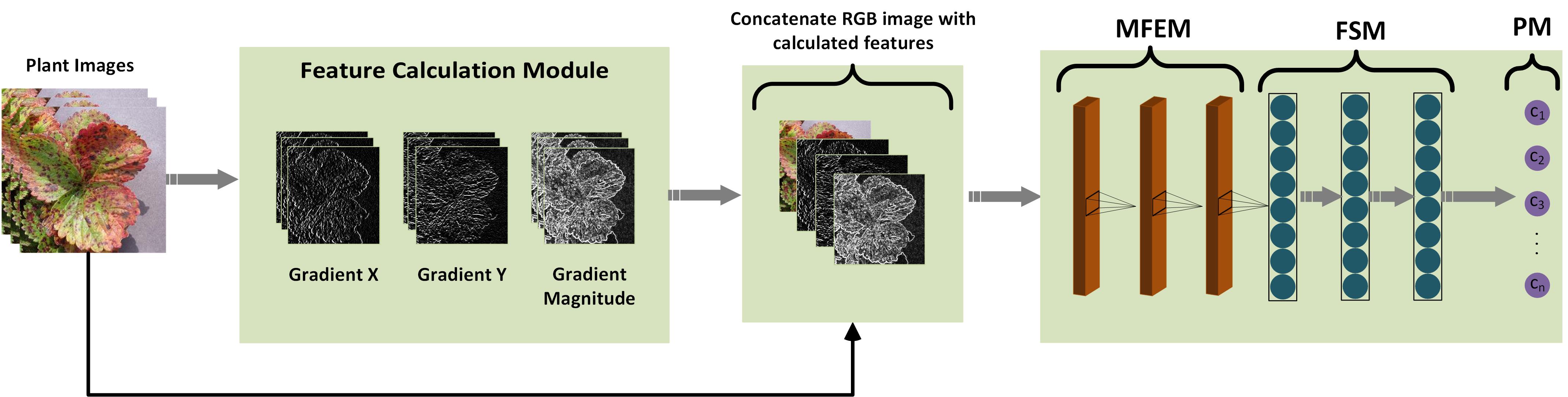}
	\caption{Diagram of the proposed MV-CNN model.}
	\label{Fig.1}
\end{figure*}

In plant disease images, the affected leaf region differs from the rest of the image. The gradients of these images highlight the affected region. So, the model's classification accuracy can be improved by providing the gradient information and the RGB image. This will improve the classification model's efficiency without requiring deeper layers. 

\subsection{Images Gradient Calculation Module}
The MV-CNN model takes the training data in the form of $(X,Y)$. Where $X$ is the input training data represented as $\mathbb{R}^{a\times b\times f}$, where $a$ and $b$ indicate the height and width, while $f$ indicates the number of feature views of the input image. $Y$ is the one-hot vector representation of the ground truth labels, where $Y \in H$ and $H$ represents the unique classes of the plant diseases.
Different features views computed from the input RGB image are the horizontal gradient $Grad_X$, vertical gradient $Grad_Y$, and the gradient magnitude $Grad_M$.

\begin{figure}[!h]
	\begin{center}
		\includegraphics[width=1\linewidth]{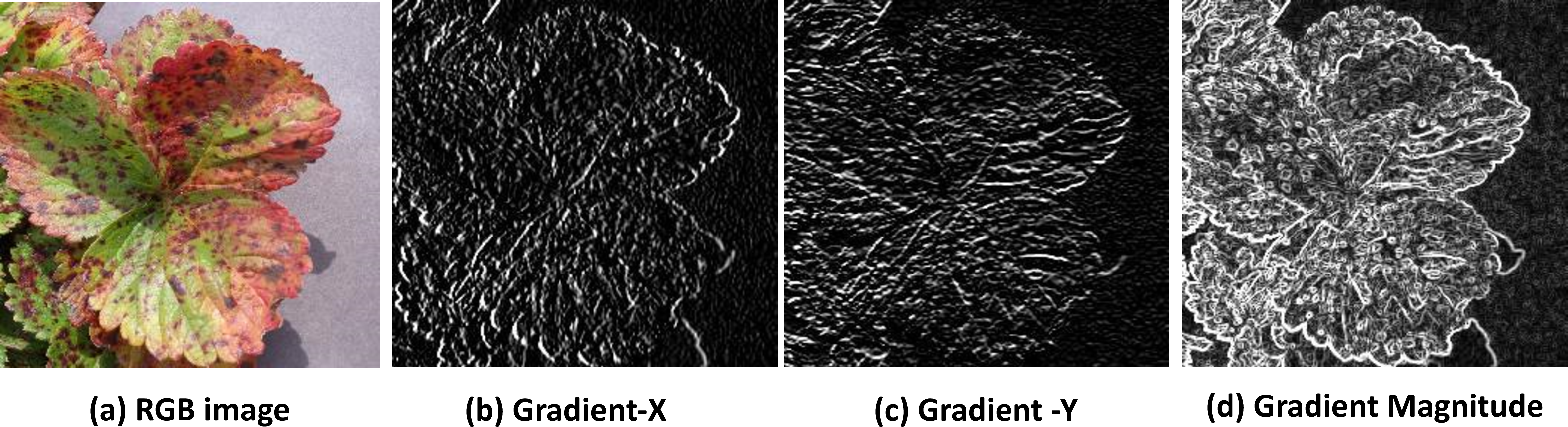}
	\end{center}
	\caption{Different feature views of plant images, created from Plantvillage \cite{Dataset} dataset. }
	\label{Fig.2}
\end{figure} 


$Grad_X$ and $Grad_Y$ are computed using Eqs.~\ref{eq:gradx} and \ref{eq:grady}, where $\Delta_x$ and $\Delta_y$ represent the gradient components along the horizontal and vertical directions of the image, respectively, computed using a Gaussian filter. The indices $i$ and $j$ denote the row and column coordinates of the pixels. Figure~2 shows a pictorial representation of these feature views.

\begin{equation}
Grad_X = \sum_{s=-d}^{d} \sum_{t=-d}^{d} \left[ \Delta_x(x_{i+s}, y_{j+t}) \right]^2.
\label{eq:gradx}
\end{equation}


\begin{equation}
Grad_Y = \sum_{s=-d}^{d} \sum_{t=-d}^{d} \left[ \Delta_x(x_{i+s}, y_{j+t}) \right]^2.
\label{eq:grady}
\end{equation}

$Grad_M$ is computed using $Grad_X$ and $Grad_Y$ as shown in Eq.~\ref{eq:grad_mag}.

\begin{equation}
Grad_M = \sqrt{Grad_X^2 + Grad_Y^2}
\label{eq:grad_mag}
\end{equation}

Given the input images $ X $, the plant disease identification model's objective is to learn a function $g$ that gives the output label $\hat{y}$ corresponding to the ground-truth label $Y$.

\begin{equation}
\hat{y} = g(H \mid R; u, v)
\label{Equ.4}
\end{equation}
where $R$ is the representation function and $u$, $v$ are its parameters. Function $g$ is parameterized with the softmax activation function.

\begin{equation}
\hat{y} = \text{softmax}(H|R; u, v).
\label{eq:Equ.5}
\end{equation}

The softmax function computes the normalized probability for each class \cite{bouchard2007efficient}.
The representation function $R$ represents the proposed MV-CNN model.

\begin{equation}
R = \text{MV-CNN}(X)
\label{eq:Equ.6}
\end{equation}

More specifically, it can be described as

\begin{multline}
\text{MV-CNN}(X) = FSM_{(N)}\cdots(FSM_{(1)}\\(MFEM_{(M)}\cdots(MFEM_{(1)}(X)))),
\label{eq:Equ.7}
\end{multline}

\subsection{Multi Feature Extraction Modules}
where $M$ represents the number of multi-feature extraction modules, while $N$ represents the number of feature selection modules. Features are extracted at each layer $l$ of the MFEM module, which are represented as $h_{(l)}$, while $MFEM_{(l)}$ indicates the extracted features. 

\begin{multline}
MFEM_{(l)} = h_{(l)}(MFEM_{(l-1)}; \omega_{(l)}; b_{(l)})
=\\ f_{norm}(\rho(f_{act}(\omega_{(l)}\ast MFEM_{(l-1)} + b_{(l)}))
\label{eq:Equ.8}
\end{multline}

where, $\omega_{(l)}$ are the weights and $b_{(l)}$ is the bias of $l^{th} $ layer while, $f_{act}$, $\ast$, and  $f_{norm}$ indicates the non-linear activation function, convolution layer operation, and normalization function respectively, and $\rho$ describes the maxpooling layer.

$MFEM_{(l-1)}$ is either the input image $X$ for the first layer $(l=1)$ or activation for the other $l-1$ layers where $(l>1)$. MFEM is the combination of convolution, activation, normalization, and max pooling functions, respectively. In the convolutional layer, the input image is convolved with filters that extract useful features and produce a feature map. The convolution layer's output is passed through a non-linear activation function (such as ReLU), followed by a batch normalization operation and max pooling. The max-pooling layer takes the maximum value from each local region and combines the locally associated features into a single feature.
\subsection{Feature Selection Module}
The MFEM output is passed to the FSM for feature selection and prediction. FSM consists of
fully connected layers in which every neuron of a layer is connected with each neuron of the previous layer. This layer computes the dot product and applies a non-linear activation function.

\begin{multline}
FSM_{(l)} = m_{(l)}(FSM_{(l-1)}; \omega_{(l)}; b_{(l)})
\\= f_{act}(\omega_{(l)} . FSM_{(l-1)} + b_{(l)}),
\label{eq:Equ.9}
\end{multline}

where $FSM_{(l-1)}$ indicates the $(l-1)^{th}$ layer activation when $(l>M)$ and $MFEM_{(M)}$ when $(l=M)$. The network parameters $f$ and $R$ are optimized with the Adam optimizer and the categorical cross-entropy loss function.
\subsection{Prediction Module}
After the FSM, these selected features are passed to the Softmax activation for the network's decision. Dropout is used for network regularization. The proposed network diagram is shown in Figure 3.

\begin{figure}[!h]
	\begin{center}
		\includegraphics[width=1\columnwidth]{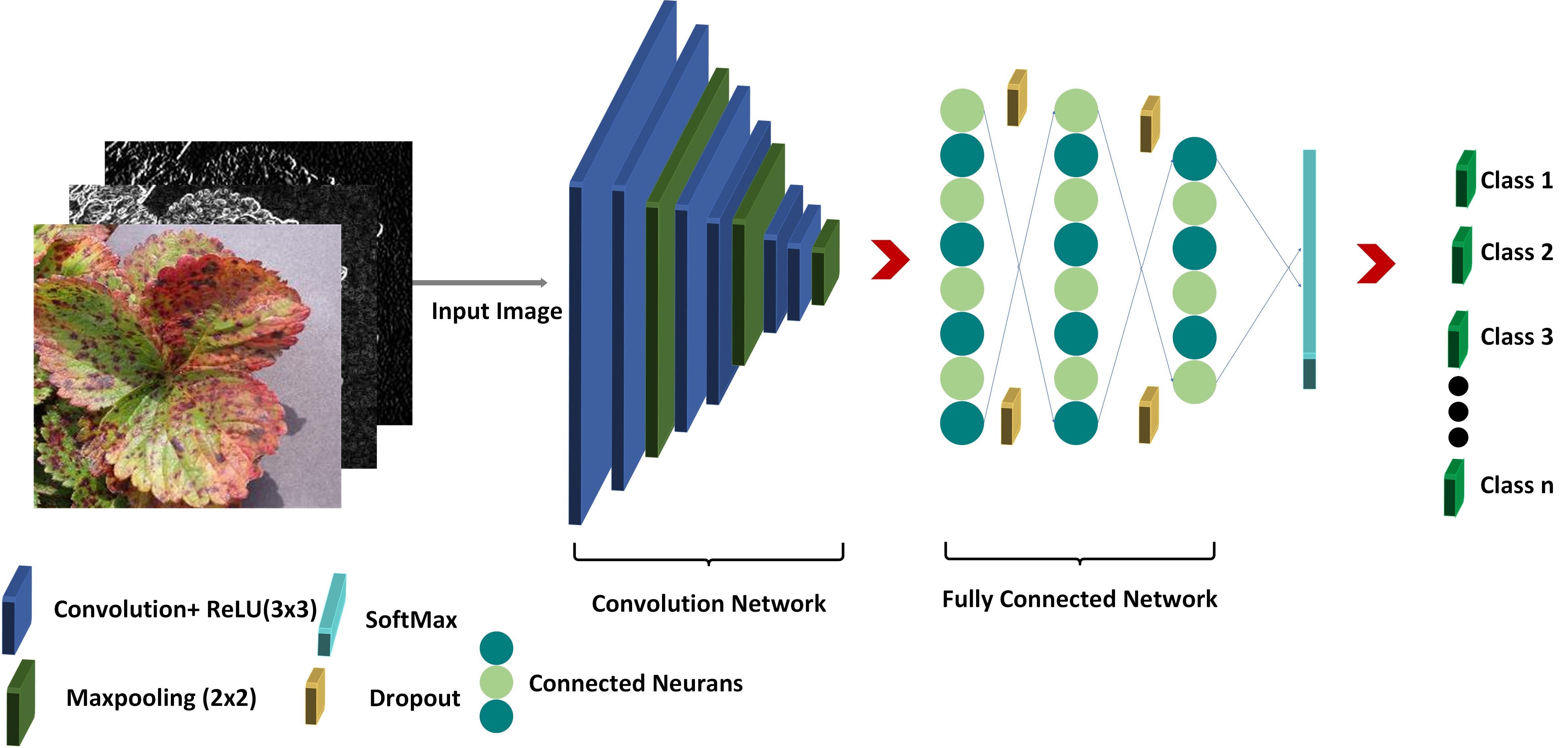}
	\end{center}
	\caption{Proposed MV-CNN model with multi-feature view input.}
	\label{Fig.3}
\end{figure} 

\section{EXPERIMENTS AND RESULTS}

The PlantVillage dataset \cite{Dataset} is used for the experimentation. This dataset consists of $54305$ color images of different plant leaves with diseases. It has 38 classes, each with 14 plant categories. This dataset is divided into $43456$ training and $10849$ validation images. The dataset is highly imbalanced, with each class having a different number of images. For example, the healthy potato class has only 122 images for training, while the tomato with yellow leaf curl virus class has 4286 images. It is a huge challenge to manage this highly imbalanced dataset.

The Google Colaboratory \cite{GoogleColab} platform with the TensorFlow \cite{abadi2016tensorflow} library is used for implementation. 

\begin{table}[]
\centering
\caption{Proposed MV-CNN model's Architecture.}
\label{Tab.1}
\resizebox{\columnwidth}{!}{%
\begin{tabular}{|c|c|c|c|c|c|}
\hline
\textbf{Layer}     & \textbf{Filter Size} & \textbf{No. Of Filters} & \textbf{Input} & \textbf{OutPut} & \textbf{Activation} \\\hline
Conv-1             & 3 × 3                & 32                      & 256 × 256 × 32 & 254 × 254 × 32  & ReLU                \\ \hline
Conv-2             & 3 × 3                & 32                      & 254 × 254 × 32 & 252 × 252 × 32  & ReLU                \\ \hline
Pool-1             & 2 × 2                & -                       & 252 × 252 × 32 & 126 × 126 × 32  & -                   \\ \hline
Conv-3             & 3 × 3                & 64                      & 126 × 126 × 32 & 124 × 124 × 64  & ReLU                \\ \hline
Conv-4             & 3 × 3                & 64                      & 124 × 124 × 64 & 122 × 122 × 64  & ReLU                \\ \hline
Pool-2             & 2 × 2                & -                       & 122 × 122 × 64 & 61 × 61 × 64    & -                   \\ \hline
Conv-5             & 3 × 3                & 128                     & 61 × 61 × 64   & 59 × 59 × 128   & ReLU                \\ \hline
Conv-6             & 3 × 3                & 128                     & 59 × 59 × 128  & 57 × 57 × 128   & ReLU                \\ \hline
Pool-3             & 2 × 2                & -                       & 57 × 57 × 128  & 28 × 28 × 128   & -                   \\ \hline
Conv-7             & 3 × 3                & 256                     & 28 × 28 × 128  & 26 × 26 × 256   & ReLU                \\ \hline
Pool-4             & 2 × 2                & -                       & 26 × 26 × 256  & 13 × 13 × 256   & -                   \\ \hline
Globel Avg Pooling &                      &                         &                & 256             & -                   \\ \hline
Dense              &                      &                         &                & 38              & SoftMax               \\  \hline          
\end{tabular}%
}
\end{table}

\subsection{NETWORK ARCHITECTURE}

Four network parameters are tested with different settings, as described below, to obtain an efficient and accurate plant disease classification model. 

\begin{itemize}
\item For the feature extraction module, different numbers of convolution layers varying between 3 and 8 are tested with max-pooling applied after each layer or after stacking 2-3 layers.
\item in convolution layers, non-linear activation function, e.g. ReLU and leaky ReLU, are applied.
\item Batch normalization layers are used after one or two convolution layers. 
\item Different optimizers, such as Adadelta, SGD, and Adam, are tested.
\item Dropout rates between 20 \% to 50 \% are evaluated after different layers.
\end{itemize}

After exhaustively evaluating different hyperparameter settings, we found that the best-performing network configuration consisted of 7 convolutional layers. The batch normalization layer improved the network's stability and accelerated convergence during training. The ReLU activation function performed better than Leaky-ReLU. The Adam optimizer demonstrated faster convergence and higher accuracy than the other two optimizers tested. The proposed MV-CNN with the best performing architecture is shown in Table 2.

\subsection{MULTI-VIEW NETWORK CONFIGURATION}
In this section, the proposed MV-CNN architecture is evaluated with different multi-view combinations. Four types of views can be used as input to assess the network: the original RGB image $Img_{RGB}$ of the plant disease, the horizontal gradient $Grad_X$ of the original image, the vertical gradient $Grad_Y$ of the original image, and the gradient magnitude $Grad_M$ of the image. The results of different feature view combinations are presented in Table 3.

Firstly, the proposed MV-CNN model is trained and tested only on RGB images $Img_{RGB}$ as input. The MV-CNN with three-channel input achieved training and test accuracies of $95.18$\% and $97.42$\%, respectively. Secondly, when $Grad_M$ is used as an additional channel alongside the RGB image, the four-channel combination ($Img_{RGB}$ + $Grad_M$) improves network performance, achieving $99.76$\% training and $99.23$\% testing accuracy. In other experiments, $Grad_X$ and $Grad_Y$ are used as additional channels alongside $Img_{RGB}$. This five-channel combination ($Img_{RGB}$+ $Grad_X$ $Grad_Y$) achieved $99.67$\% and $98.36$\% training and testing accuracy, respectively. When all multi-view features are used together with the RGB image ($Img_{RGB}$+ $Grad_X$ $Grad_Y$ $Grad_M$), the six-channel combination achieved $99.68$\% training and $99.07$\% testing accuracy. These results show that additional information improves the network's performance. Among the tested combinations, the four-channel combination of $Img_{RGB}$ + $Grad_M$ achieved the best accuracy. The training, validation, accuracy, and loss graphs for the proposed $Img_{RGB}$ + $Grad_M$ are presented in Figure 4.

\begin{figure}[!h]
	\begin{center}
		\includegraphics[width=0.7\linewidth]{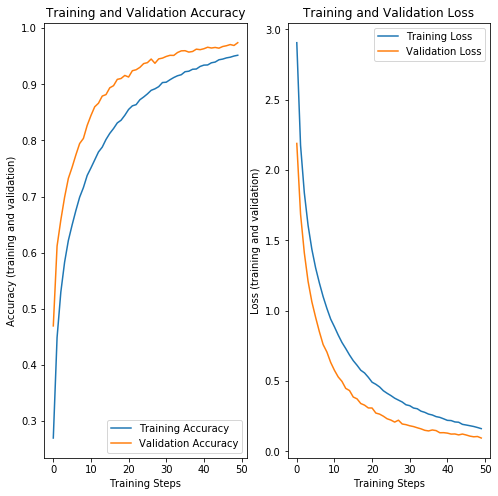}
	\end{center}
	\caption{Training and validation Accuracy graph of the proposed Model.}
	\label{fig:1}
\end{figure}

\par

Additionally, the results of different feature combinations are also shown in Figure 4. These feature maps are generated using the Grad-CAM localization Method \cite{selvaraju2017grad}, which highlights the important features on the plant leaf that the network found helpful for classification. It shows a visualization of the same leaf images at the last layers of the network, indicating that the ($Img_{RGB}$ + $Grad_M$) combination produces more accurate feature maps than the other combinations.

\begin{figure}[!h]
	\begin{center}
		\includegraphics[width=1\linewidth]{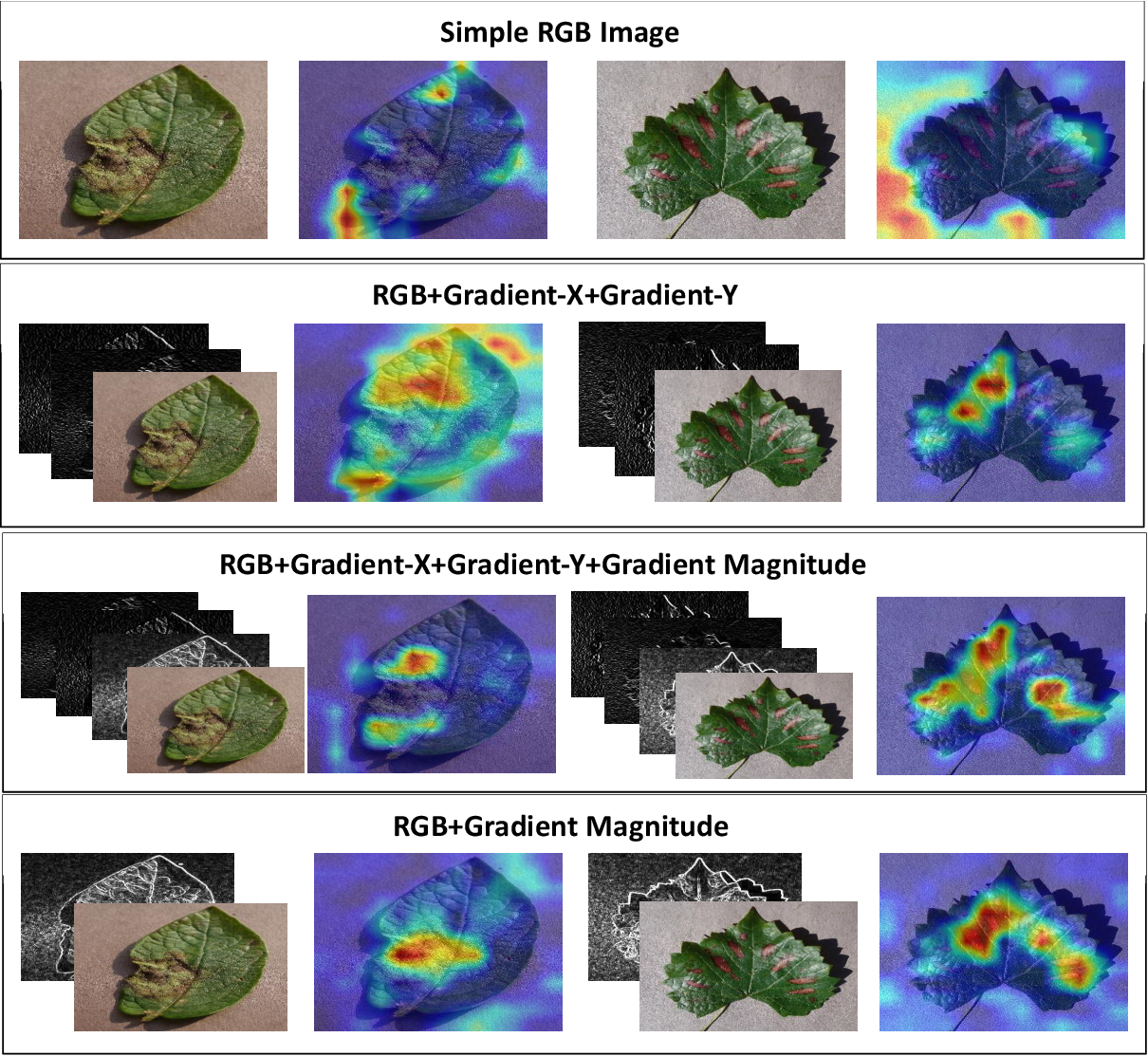}
	\end{center}
	\caption{Features map of the last layer of the proposed MV-CNN model with different feature views combination of plant images of the Plantvillage \cite{Dataset} dataset.}
	\label{fig:1}
\end{figure}

In the same way, ($Img_{RGB}$+ $Grad_X$+ $Grad_Y$) and ($Img_{RGB}$+ $Grad_X$+ $Grad_Y$+$Grad_M$) combinations also demonstrate useful feature maps as compared to the simple RGB image. Hence, pre-calculated features and the RGB image have been shown to be more useful than a simple RGB image alone.

\begin{table}[]
\centering
\caption{Proposed Models Accuracy with Different Data Features Combinations on Plantvillage \cite{Dataset} DataSet}
\label{Tab.2}
\resizebox{\columnwidth}{!}{%
\begin{tabular}{|c|c|c|c|}

\hline
\textbf{MV\_CNN Model} & \textbf{Layers} & \textbf{Training Acc.} & \textbf{Validation Acc.} \\\hline
\textbf{$Img_{RGB}$}                         & 7 & 95.18          & 97.42          \\\hline
\textbf{$Img_RGB+Grad_X+Grad_Y$}         & 7 & 99.67          & 98.36          \\\hline
\textbf{$Img_RGB+Grad_X+Grad_Y+Grad_M$} & 7 & 99.68          & 99.07          \\\hline
\textbf{$Img_RGB+Grad_M$}                 & 7 & \textbf{99.76} & \textbf{99.36} 
\\\hline
\end{tabular}%
}
\end{table}

\subsection{NETWORK PERFORMANCES}
Two different types of performance analysis are presented in this section. First, the comparison of the various models trained from scratch is discussed. Secondly, the comparison results of models pre-trained on the ImageNet dataset are discussed. The model comparisons are presented in terms of network layers and complexity, training time, and number of network parameters.

\begin{table}[]
\centering
\caption{Proposed MV-CNN Models Accuracy and layers Comparison with baseline networks on Plantvillage \cite{Dataset} benchmark dataset.}
\label{Tab.3}
\begin{tabular}{|c|c|c|}

\hline
\textbf{Models} & \textbf{Layers} & \textbf{Accuracy} 
\\\hline
{CNN\cite{geetharamani2019identification}} & 9                                   & 96.46                                 \\\hline
{CNN\cite{kaya2019analysis}}              & 15                                  & 97.40                                 \\\hline
{CNN-RNN\cite{kaya2019analysis}}          & {-}               & 98.77                                 \\\hline
\textbf{Proposed ($Img_{RGB}+Grad_M$)}                 & \textbf{7} & \textbf{99.36} 
\\\hline
\end{tabular}
\end{table}

Table 4 presents the network performance of models trained from scratch. The performance of the proposed network is compared with \cite{geetharamani2019identification} and \cite{kaya2019analysis}.
It can be seen that the model in \cite{geetharamani2019identification} has a nine-layer architecture and it achieves an accuracy of $96.46$\%. Kaya \textit{et al.}\cite{kaya2019analysis} proposed two different networks, a conventional CNN and a CNN-RNN model. The conventional CNN model has a 15-layer architecture and achieves an accuracy of $97.40$\%. When the CNN model is combined with the RNN, it achieves $98.77$\% accuracy. On the other hand, our proposed model with only seven layers achieved $99.36$\% accuracy, outperforming the models mentioned above. The accuracy comparison between the baseline model and the proposed MV-CNN model, with different feature view combinations, on the Plantvillage \cite{Dataset} dataset is presented in Figure 6.


\begin{figure}
  \centering
  \includegraphics[width=7cm]{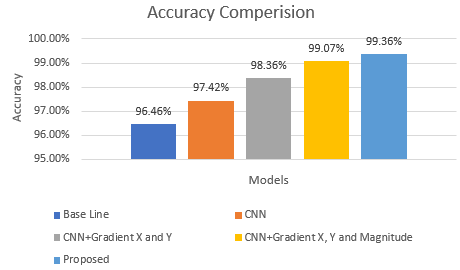}
  \caption{Accuracy Comparison of Base Line Model with proposed MV-CNN model with different feature views combination of Plantvillage \cite{Dataset} dataset.}
  \label{fig:1}
\end{figure}

Table 5 presents a comparison of pre-trained models. \cite{kaya2019analysis,too2019comparative} performed experiments on Plantvillage dataset \cite{Dataset} using SOTA networks. Too \textit{et al.}\cite{too2019comparative} used Inception-V4, VGGNet, ResNet-50, ResNet-101, and DenseNet, while Kaya \textit{et al.}\cite{kaya2019analysis} performed experiments using AlexNet. Inception-V4 has 41.3 million parameters, and it achieved 98.02\% accuracy with an average training time of 4042 seconds per epoch. The two ResNet architectures with 50 and 101 layers, and 23.6 and 42.5 million parameters, took 1583 and 2766 seconds per epoch, respectively, and achieved $99.67$\% and $99.66$\% accuracy. VGGNet with 16 layers and 119.6 million parameters achieved $81.92$\% accuracy in 1051 seconds per epoch. DenseNet with 121 layers and 7.1 million parameters took 2165 seconds per epoch and achieved $99.76$\% accuracy. Finally, the proposed model with only 7 layers and 0.726 million parameters, achieved a maximum of $99.36$\% accuracy in only 337 seconds per epoch. These results demonstrate that the proposed model achieved highly comparable performance to SOTA models while being much less computationally expensive.

\begin{table}[]
\centering
\caption{Proposed MV-CNN models comparison with some SOTA pre-trained networks in terms of network layers, number of parameters and training time per epoch.}
\label{tab:my-table}
\resizebox{\columnwidth}{!}{%
\begin{tabular}{|c|c|c|c|c|}
\hline
\textbf{Models} & \textbf{Layers} & \textbf{Params} & \textbf{Acc (\%)} &\textbf{Time (sec)} \\\hline
\textbf{Inception V4\cite{too2019comparative}} & -                                  & 41.2M & 98.02 & 4042                                 \\\hline
\textbf{VGG Net\cite{too2019comparative}} & 16                                  & 119.6M & 81.92 & 1051                                   \\\hline
\textbf{ResNet 50\cite{too2019comparative}} & 50                                  & 23.6M & 99.67 & 1583                                  \\\hline
\textbf{ResNet 101\cite{too2019comparative}} & 101                                 & 42.5M & 99.66 & 2766                                  \\\hline
\textbf{DenseNet\cite{too2019comparative}} & 121                                 & 7.1M & 99.76 & 2165                                  \\\hline
\textbf{AlexNet \cite{kaya2019analysis}} & -                                 & - & 98.60 & -                                  \\\hline
\textbf{Proposed (MV-CNN)}                 & \textbf{7} & \textbf{0.726M} & 99.36 & \textbf{337}
\\\hline
\end{tabular}%
}
\end{table}

In addition, the proposed MV-CNN model is evaluated using the confusion matrix, the Area Under the Receiver Operating Characteristic (AUC–ROC) curve \cite{narkhede2018understanding}, and other metrics, including precision, recall, and F1 score.
The confusion matrix shows the performance of the proposed model on test data against known true labels. It can also be called a matching matrix, where each row represents the predicted values and each column shows the actual value. Appendix A represents the confusion matrix of the proposed MV-CNN classification model, which describes the accurately predicted classes against each class

The AUC metric is one of the most popular methods for evaluating network performance. The ROC Curve is used to show the relationship between the True Positive (TP) rate and the False Negative (FN) rate. FP rate represents the number of incorrectly predicted negative data points that are predicted as positive relative to all negatives, and is calculated using equation 10.

\begin{equation}
FP_{Rate} = \frac{FP}{TN+FP}    
\end{equation}

TP rate indicates the number of correctly predicted positive data points, which are predicted as negative with respect to all positive, and calculated as   
\begin{equation}
TP_{Rate} = \frac{TP}{FN+TP}
\end{equation}

In addition, the Class-wise detailed comparison of ROC curves of the simple CNN model with $Img_{RGB}$ input, the proposed MV-CNN model with $Img_{RGB}+Grad_M$ input and the ROC curves of the SOTA ResNet50 model \cite{too2019comparative} with $Img_{RGB}$ input. This describes that the proposed model with four-channel input performs better than the simple CNN model with only RGB input. These AUC-ROC curves for each class also show that the proposed model yields highly comparable results to the ResNet50 \cite{too2019comparative} model.

The difference between the AUC-ROC curves and the accuracy of these three models is presented in Appendix B, which shows that the proposed model's overall performance is better than that of the Simple CNN model. These AUC-ROC curves and Accuracy also explain the difference between the proposed model and the SOTA ResNet50 \cite{too2019comparative} model. The performance of the proposed model and ResNet50 is the same across most classes, including healthy ones. The proposed model yields results that are highly comparable to those of the ResNet50 \cite{too2019comparative} model. The difference between the AUC-ROC curves and the accuracy of these three models is presented in Table 5.

\begin{table}[]
\centering
\caption{Detail comparison of AUC-ROC Curves of simple CNN model with $Img_{RGB}$ input, proposed MV-CNN model with  $Img_{RGB}+Grad_M$ input and  SOTA ResNet50 model \cite{too2019comparative} with $Img_{RGB}$.}
\label{tab:my-table}
\resizebox{\columnwidth}{!}{%
\begin{tabular}{|c|c|c|c|}
\hline
\textbf{Class Name}                     & \textbf{Proposed RGB} & \textbf{Proposed MV} & \textbf{ResNet 50}  \\\hline
\text{Apple-scab}                & \text{0.9994}      & \text{1.0000}   & \text{1.0000}     \\\hline
\text{Apple-blackrot}           & \text{1.0000}      & \text{1.0000}   & \text{1.0000}       \\\hline
\text{Apple-cedar apple rust}    & \text{1.0000}      & \text{1.0000}   & \text{1.0000}       \\\hline
\text{Healthy-apple}                  & \text{0.9998}      & \text{1.0000}   & \text{1.0000}      \\\hline
\text{Blueberry-healthy}         & \text{0.9999}      & \text{1.0000}   & \text{1.0000}    \\\hline
\text{Cherry-powdery mildew}     & \text{1.0000}      & \text{1.0000}   & \text{1.0000}     \\\hline
\text{Cherry-healthy}            & \text{0.9998}      & \text{0.9985}   & \text{0.9961}     \\\hline
\text{Peach-bacterial spot}      & \text{0.9999}      & \text{0.9999}   & \text{1.0000}     \\\hline
\text{Corn-grey leaf spot}       & \text{0.9991}      & \text{0.9996}   & \text{0.9999}     \\\hline
\text{Grape-black rot}           & \text{0.9997}      & \text{1.0000}   & \text{1.0000}       \\\hline
\text{Corn-northern leaf blight} & \text{0.9993}      & \text{0.9995}   & \text{0.9999}      \\\hline
\text{Corn-common rust}          & \text{1.0000}      & \text{1.0000}   & \text{1.0000}      \\\hline
\text{Healthy-corn}                   & \text{1.0000}      & \text{1.0000}   & \text{1.0000}       \\\hline
\text{Grape-black measles}       & \text{0.9999}      & \text{1.0000}   & \text{1.0000}       \\\hline
\text{Grape-leaf blight}         & \text{1.0000}      & \text{1.0000}   & \text{1.0000}       \\\hline
\text{Healthy-grape}                  & \text{1.0000}      & \text{1.0000}   & \text{1.0000}       \\\hline
\text{Orange-Huanglongbing}      & \text{1.0000}      & \text{1.0000}   & \text{1.0000}     \\\hline
\text{Healthy-peach}                  & \text{1.0000}      & \text{1.0000}   & \text{1.0000}      \\\hline
\text{Healthy-pepper}                 & \text{1.0000}      & \text{1.0000}   & \text{1.0000}     \\\hline
\text{Potato-early blight}       & \text{0.9998}      & \text{1.0000}   & \text{1.0000}     \\\hline
\text{Healthy-potato}                 & \text{0.9999}      & \text{1.0000}   & \text{1.0000}     \\\hline
\text{Healthy-raspberry}              & \text{0.9999}      & \text{1.0000}   & \text{1.0000}      \\\hline
\text{Healthy-soybean}                & \text{0.9999}      & \text{1.0000}   & \text{1.0000}      \\\hline
\text{Squash-powdery mildew}     & \text{1.0000}      & \text{1.0000}   & \text{1.0000}    \\\hline
\text{Healthy-strawberry}             & \text{0.9999}      & \text{1.0000}   & \text{1.0000}      \\\hline
\text{Strawberry-leaf scorch}    & \text{0.9999}      & \text{1.0000}   & \text{1.0000}    \\\hline
\text{Tomato-bacterial spot}     & \text{0.9999}      & \text{1.0000}   & \text{1.0000}    \\\hline
\text{Tomato-early blight}       & \text{0.9995}      & \text{0.9998}   & \text{1.0000}     \\\hline
\text{Tomato-late blight}                 & \text{0.9993}      & \text{0.9999}   & \text{1.0000}      \\\hline
\text{Tomato-leaf mold}        & \text{0.9997}      & \text{1.0000}   & \text{1.0000}    \\\hline
\text{Tomato-septoria leaf spot}          & \text{0.9996}      & \text{0.9996}   & \text{1.0000}      \\\hline
\text{Pepper-bacterial spot}     & \text{0.9998}      & \text{1.0000}   & \text{1.0000}     \\\hline
\text{Tomato-two spotted spider mite} & \text{0.9981}      & \text{0.9998}   & \text{1.0000}      \\\hline
\text{Tomato-target spot} & \text{0.9997} & \text{0.9997} & \text{0.9999}  \\\hline
\text{Potato-late blight}        & \text{0.9997}      & \text{1.0000}   & \text{1.0000}      \\\hline
\text{Tomato-mosaic virus}        & \text{0.9999}      & \text{1.0000}   & \text{1.0000}        \\\hline
\text{Tomato-yellow leaf curl virus}       & \text{0.9999}      & \text{1.0000}   & \text{1.0000}      \\\hline
\text{Healthy-tomato}  & \text{0.9999} & \text{1.0000} & \text{1.0000}  \\\hline
\end{tabular}%
}
\end{table}

Table 5 shows that the proposed model's performance is overall better than that of the Simple CNN model. These AUC-ROC curves and Accuracy also explain the difference between the proposed model and the SOTA ResNet50 \cite{too2019comparative} model. The performance of the proposed model and ResNet50 is the same across most classes, including healthy ones. The proposed model yields results that are highly comparable to those of the ResNet50 \cite{too2019comparative} model.            

Table 6 presents the precision, recall and F1 score of the proposed model. Precision is calculated by dividing the TP predictions by the total number of positive predicted values $(FP+TP)$ \cite{shung2018accuracy}. The precision value is calculated using the equation below.
A precision comparison of the baseline model and the proposed MV-CNN model, with different feature view combinations, on the Plantvillage \cite{Dataset} dataset, is presented in Figure 7.

\begin{figure}[!h]
	\begin{center}
		\includegraphics[width=1\linewidth]{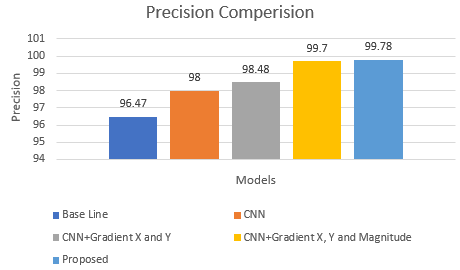}
	\end{center}
	\caption{Precision Comparison of Base Line Model with proposed MV-CNN model with different feature views combination of Plantvillage \cite{Dataset} dataset.}
	\label{fig:1}
\end{figure}

\begin{equation}
P= \frac{TP}{FP+TP}
\end{equation}
Recall \cite{kynkaanniemi2019improved} is calculated by dividing TP values by the sum of TP and FN values. A detailed comparison of Recall is presented in Figure 8 between the baseline model and the proposed MV-CNN model, with different feature view combinations of the PlantVillage \cite{Dataset} dataset, as shownThe  in Figure 6.

\begin{figure}[!h]
	\begin{center}
		\includegraphics[width=1\linewidth]{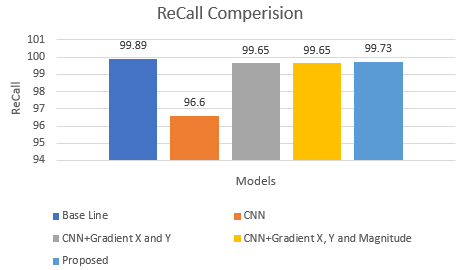}
	\end{center}
	\caption{Accuracy Comparison of Base Line Model with proposed MV-CNN model with different feature views combination of Plantvillage \cite{Dataset} dataset.}
	\label{fig:1}
\end{figure}
\begin{equation}
R= \frac{TP}{FN+TP}
\end{equation}
F1 score is the weighted average of precision and recall.  When the F1 score is $1$, it is the best value, and when it is $0$, it is the worst value. It is calculated by using equation 14. F1 Score comparison of baseline model with proposed MV-CNN model with different feature views combination of Plantvillage \cite{Dataset} dataset is presented in Figure 9.

\begin{figure}[!h]
	\begin{center}
		\includegraphics[width=1\linewidth]{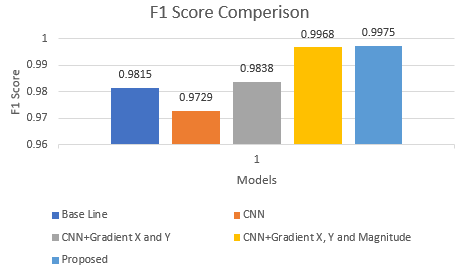}
	\end{center}
	\caption{F1 Score Comparison of Base Line Model with proposed MV-CNN model with different feature views combination of Plantvillage \cite{Dataset} dataset.}
	\label{fig:1}
\end{figure}

\begin{equation}
F1_{score}= 2*\frac{P*R}{P+R}
\end{equation}

\begin{table}[]
\centering
\caption{Precision, Recall,  F1 Score and Accuracy for each class on Plantvillage \cite{Dataset} dataset on Proposed MV-CNN is presented }
\label{tab:my-table}
\resizebox{\columnwidth}{!}{%
\begin{tabular}{|c|c|c|c|c|}
\hline
\textbf{Class Name}                     & \textbf{Precision} & \textbf{ReCall} & \textbf{F1 Score} & \textbf{AUC (\%)} \\\hline
\text{Apple-scab}                & \text{1.00}      & \text{0.94}   & \text{0.97}     & \text{1.0000}   \\\hline
\text{Apple-blackrot}           & \text{0.99}      & \text{0.99}   & \text{0.99}     & \text{1.0000}   \\\hline
\text{Apple-cedar apple rust}    & \text{0.96}      & \text{1.00}   & \text{0.98}     & \text{1.0000}   \\\hline
\text{Healthy-apple}                  & \text{1.00}      & \text{0.99}   & \text{1.00}     & \text{1.0000}   \\\hline
\text{Blueberry-healthy}         & \text{1.00}      & \text{1.00}   & \text{1.00}     & \text{1.0000}   \\\hline
\text{Cherry-powdery mildew}     & \text{0.96}      & \text{1.00}   & \text{0.98}     & \text{1.0000}   \\\hline
\text{Cherry-healthy}            & \text{0.98}      & \text{0.99}   & \text{0.99}     & \text{0.9985}   \\\hline
\text{Corn-grey leaf spot}       & \text{0.90}      & \text{0.93}   & \text{0.92}     & \text{0.9996}   \\\hline
\text{Corn-common rust}          & \text{0.99}      & \text{1.00}   & \text{0.99}     & \text{1.0000}   \\\hline
\text{Corn-northern leaf blight} & \text{0.98}      & \text{0.94}   & \text{0.96}     & \text{1.0000}   \\\hline
\text{Healthy-corn}                   & \text{1.00}      & \text{0.99}   & \text{1.00}     & \text{1.0000}   \\\hline
\text{Grape-black rot}           & \text{1.00}      & \text{0.99}   & \text{0.99}     & \text{1.0000}   \\\hline
\text{Grape-black measles}       & \text{0.99}      & \text{0.99}   & \text{0.99}     & \text{1.0000}   \\\hline
\text{Grape-leaf blight}         & \text{1.00}      & \text{1.00}   & \text{1.00}     & \text{1.0000}   \\\hline
\text{Healthy-grape}                  & \text{1.00}      & \text{0.89}   & \text{0.94}     & \text{1.0000}   \\\hline
\text{Orange-Huanglongbing}      & \text{1.00}      & \text{1.00}   & \text{1.00}     & \text{1.0000}   \\\hline
\text{Peach-bacterial spot}      & \text{1.00}      & \text{0.98}   & \text{0.99}     & \text{0.9999}   \\\hline
\text{Healthy-peach}                  & \text{0.94}      & \text{1.00}   & \text{0.97}     & \text{1.0000}   \\\hline
\text{Pepper-bacterial spot}     & \text{0.94}      & \text{0.99}   & \text{0.97}     & \text{1.0000}   \\\hline
\text{Healthy-pepper}                 & \text{0.99}      & \text{1.00}   & \text{0.99}     & \text{1.0000}   \\\hline
\text{Potato-early blight}       & \text{0.98}      & \text{0.99}   & \text{0.99}     & \text{1.0000}   \\\hline
\text{Healthy-potato}                 & \text{0.99}      & \text{0.97}   & \text{0.98}     & \text{1.0000}   \\\hline
\text{Potato-late blight}        & \text{0.87}      & \text{0.90}   & \text{0.89}     & \text{1.0000}   \\\hline
\text{Healthy-raspberry}              & \text{0.95}      & \text{0.99}   & \text{0.97}     & \text{1.0000}   \\\hline
\text{Healthy-soybean}                & \text{0.99}      & \text{1.00}   & \text{1.00}     & \text{1.0000}   \\\hline
\text{Squash-powdery mildew}     & \text{1.00}      & \text{0.99}   & \text{1.00}     & \text{1.0000}   \\\hline
\text{Healthy-strawberry}             & \text{1.00}      & \text{0.96}   & \text{0.98}     & \text{0.9999}   \\\hline
\text{Strawberry-leaf scorch}    & \text{1.00}      & \text{0.98}   & \text{0.99}     & \text{1.0000}   \\\hline
\text{Tomato-bacterial spot}     & \text{1.00}      & \text{0.99}   & \text{0.99}     & \text{1.0000}   \\\hline
\text{Tomato-early blight}       & \text{0.95}      & \text{0.95}   & \text{0.95}     & \text{0.9998}   \\\hline
\text{Tomato-late blight}                 & \text{0.98}      & \text{0.96}   & \text{0.97}     & \text{0.9999}   \\\hline
\text{Tomato-leaf mold}        & \text{1.00}      & \text{0.97}   & \text{0.99}     & \text{1.0000}   \\\hline
\text{Tomato-septoria leaf spot}          & \text{0.95}      & \text{1.00}   & \text{0.98}     & \text{0.9999}   \\\hline
\text{Tomato-two spotted spider mite} & \text{0.99}      & \text{0.94}   & \text{0.96}     & \text{0.9998}   \\\hline
\text{Tomato-target spot} & \text{0.93} & \text{0.98} & \text{0.95} & \text{0.9997} \\\hline
\text{Tomato-mosaic virus}        & \text{1.00}      & \text{1.00}   & \text{1.00}     & \text{1.0000}   \\\hline
\text{Tomato-yellow leaf curl virus}       & \text{0.99}      & \text{0.96}   & \text{0.97}     & \text{0.9999}   \\\hline
\text{Healthy-tomato}  & \text{0.99} & \text{1.00} & \text{1.00} & \text{1.0000} \\\hline
\end{tabular}%
}
\end{table}

To further demonstrate the proposed model's performance, the following subsections present a detailed category-wise analysis of the 14 crop types and their diseases. Furthermore, category-wise heat maps of the proposed model serve as visualizations of the network's decisions.  

.
\subsubsection{Apple}
This is the first category of the Plantvillage dataset with three affected and one healthy class. Heatmaps of sample leaves from the Apple category are shown in Figure 10. According to this Figure, the heatmap identifies the most affected part of the leaf, while the healthy class has no affected region.\par

\begin{figure}
  \centering
  \includegraphics[width=7cm]{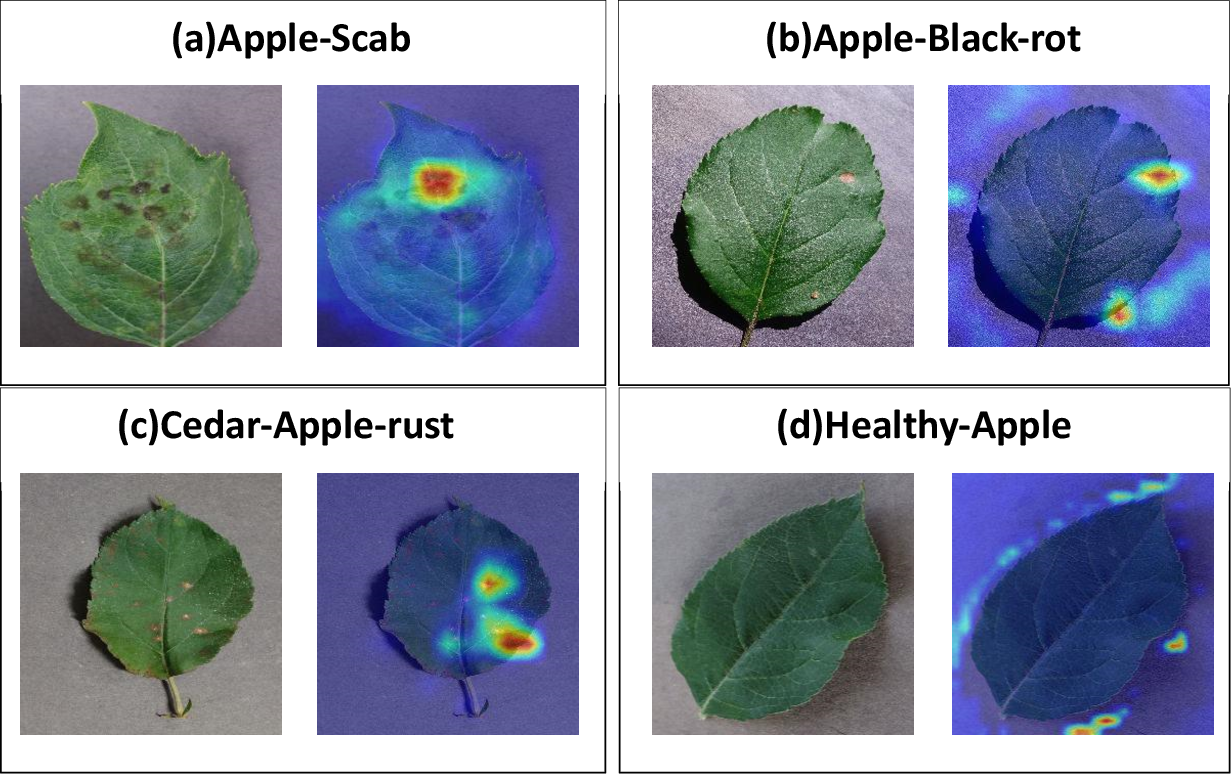}
  \captionsetup{justification=centering}
  \caption{Input images with a feature map of apple classes}
\end{figure}

\subsubsection{Blueberry}
Blueberries are the second category in the dataset, with only one class of healthy blueberries. The proposed model achieves 100\% AUC accuracy in this class. The heatmap for this class is shown in Figure 11, which highlights the area outside the leaf, indicating that there is no disease in the leaf.

\begin{figure}
  \centering
  \includegraphics[width=7cm]{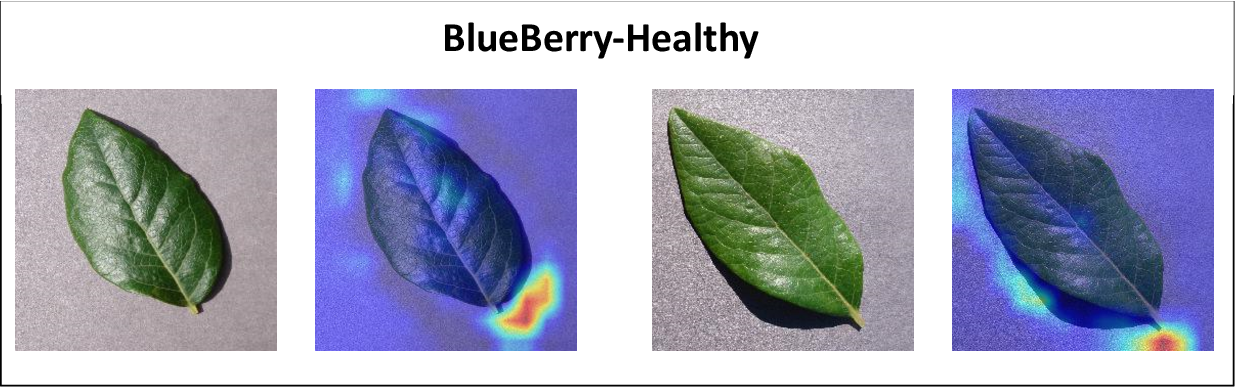}
  \caption{Features map of Blueberry healthy class }
\end{figure}

\subsubsection{Cherry}
The third category is Cherry, with two classes: healthy and diseased. Cherry-powdery-mildew has 100\% classification accuracy, while the healthy class has 99.85\% accuracy. Heatmaps for sample images are shown in Figure 12, highlighting the affected region in the diseased samples, while the healthy class shows no affected region on the leaf.

\begin{figure}
  \centering
  \includegraphics[width=7cm]{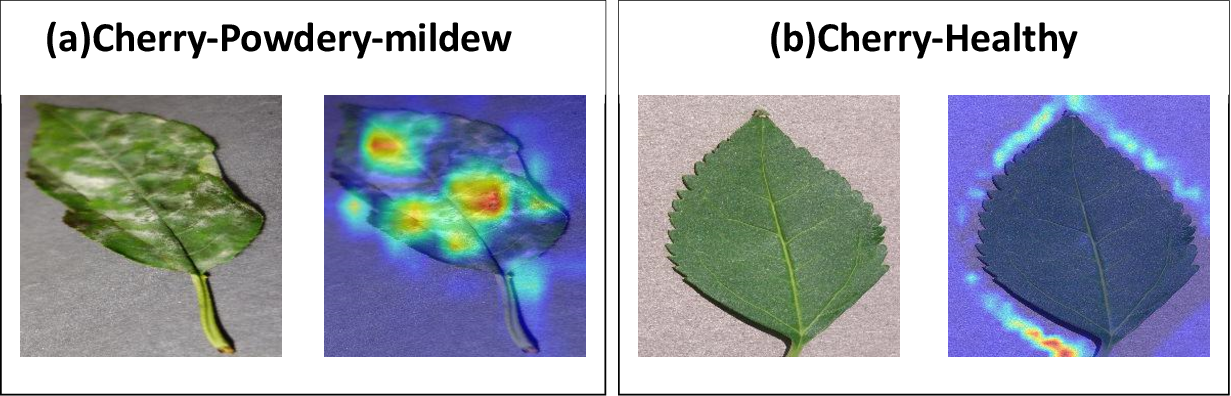}
  \caption{Input images with their feature map of cherry class, including cherry healthy  }
\end{figure}

\subsubsection{Corn (Maize)}
This plant category has four classes, including a healthy class. The diseases of the corn plant are Corn-Cercospora-leaf-spot or Gray-leaf-spot, Corn-Common-rust, Corn-Northern-Leaf-Blight and Corn-healthy. Heatmaps of these diseases are presented in Figure 13, showing the disease on the plant leaf, while the healthy class shows no disease.

\begin{figure}
  \centering
  \includegraphics[width=7cm]{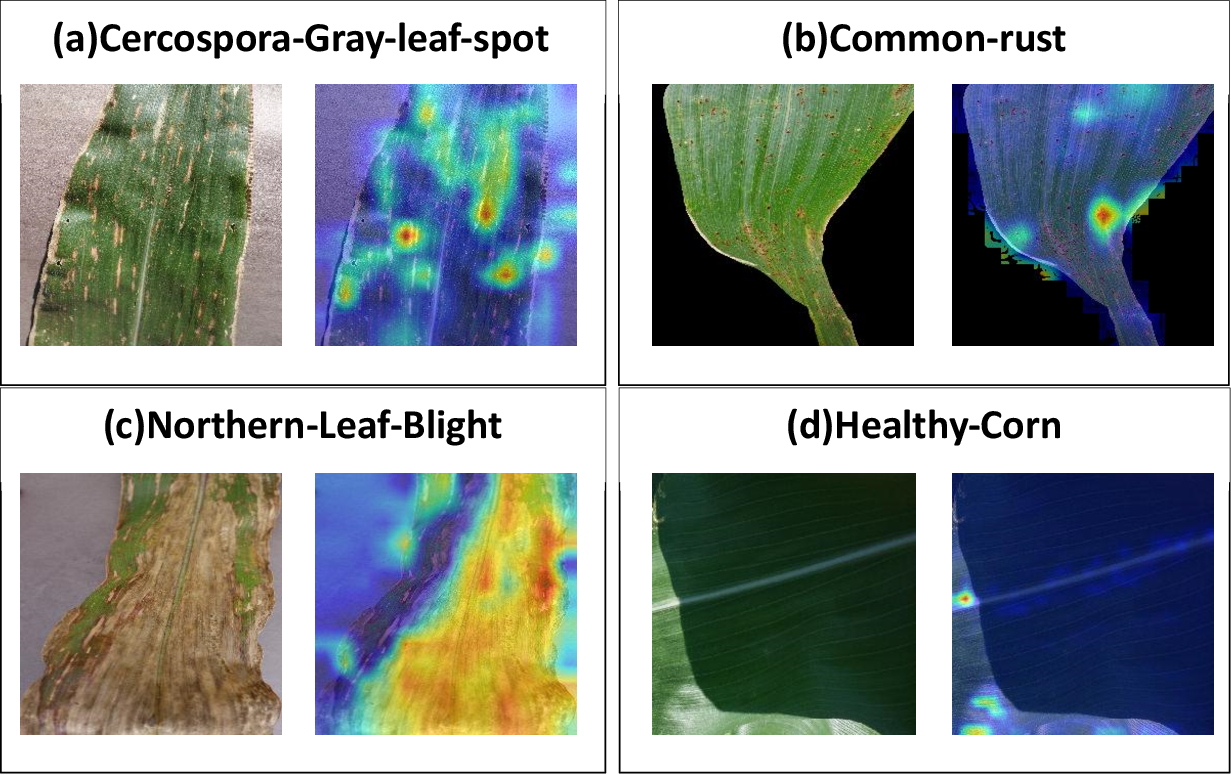}
  \caption{Features map with input images of corn}
\end{figure}

\subsubsection{Grape}
Grape is another plant category, disease classes include Grape-Black-rot, Grape-Esca-(Black-Measles) and Grape-Leaf-blight-(Isariopsis-Leaf-Spot). Diseases of grapes in the proposed model are explained using heatmaps for each class in Figure 14.

\begin{figure}
  \centering
  \includegraphics[width=7cm]{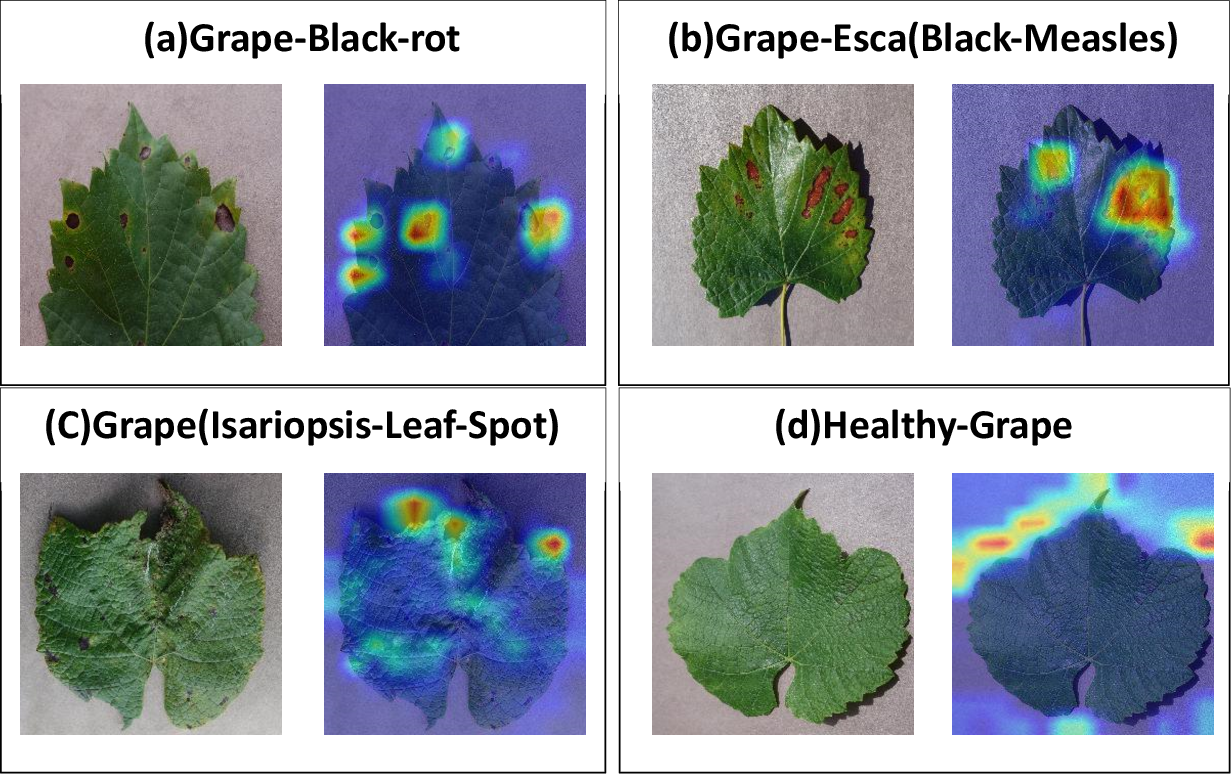}
  \caption{These are the images of Grape with their obtained features map}
\end{figure}

\subsubsection{Orange}

Orange is the sixth category, with only one affected class, Orange-Haunglongbing-(Citrus-greening). The proposed model achieves 100\% AUC accuracy on this class. Heatmaps of a sample leaf from this disease are shown in Figure 15, indicating the presence of the disease on the leaf.
\begin{figure}
  \centering
  \includegraphics[width=7cm]{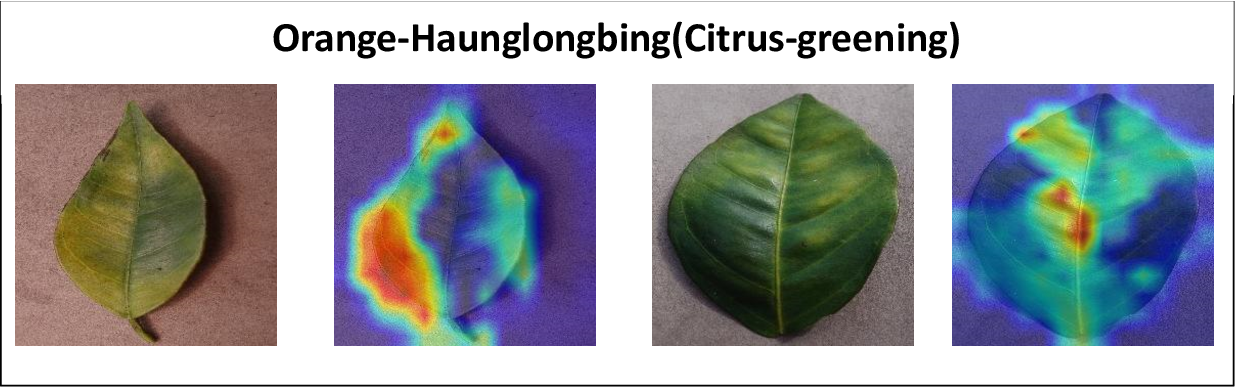}
  \caption{There is one affected class of orange with Huanglongbing, input images from this class and a features map }
\end{figure}

\subsubsection{Peach}

The peach plant category has only two classes: Peach-Bacterial-spot and Peach-healthy. Peach-Bacterial-spot gains 99.99\% accuracy, while Peach-healthy achieves 100\% accuracy. The heatmaps shown in Figure 16 show the affected leaf and the healthy class, with no affected region on the leaf.

\begin{figure}
  \centering
  \includegraphics[width=7cm]{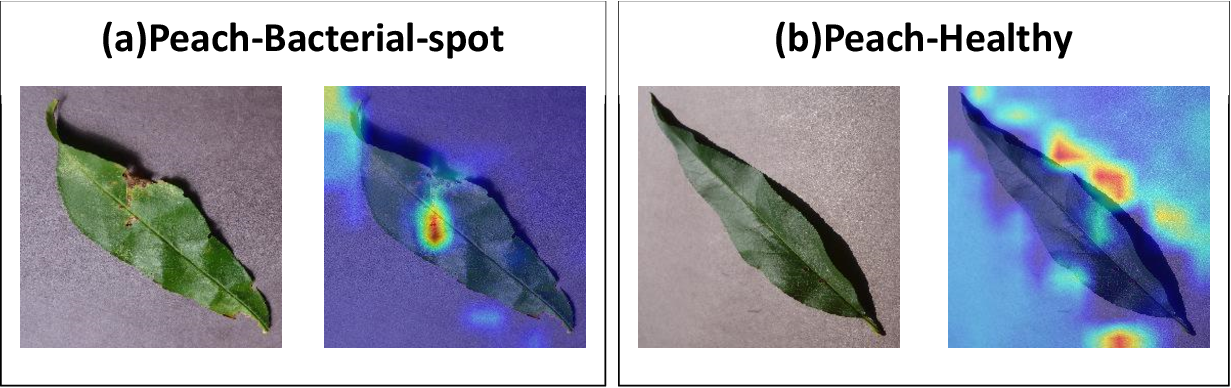}
  \caption{Input images of peach classes with their calculated feature map}
\end{figure}

\subsubsection{Pepper}

The next category is pepper that includes one Pepper-bell-Bacterial-spot and one Pepper-bell-healthy class. The proposed MV-CNN achieves 100\% accuracy across both classes. Heatmaps of samples from these two classes are shown in Figure 17. The proposed model accurately highlights the Pepper-bell-Bacterial-spot on the leaf, while the healthy leaf does not indicate disease.

\begin{figure}
  \centering
  \includegraphics[width=7cm]{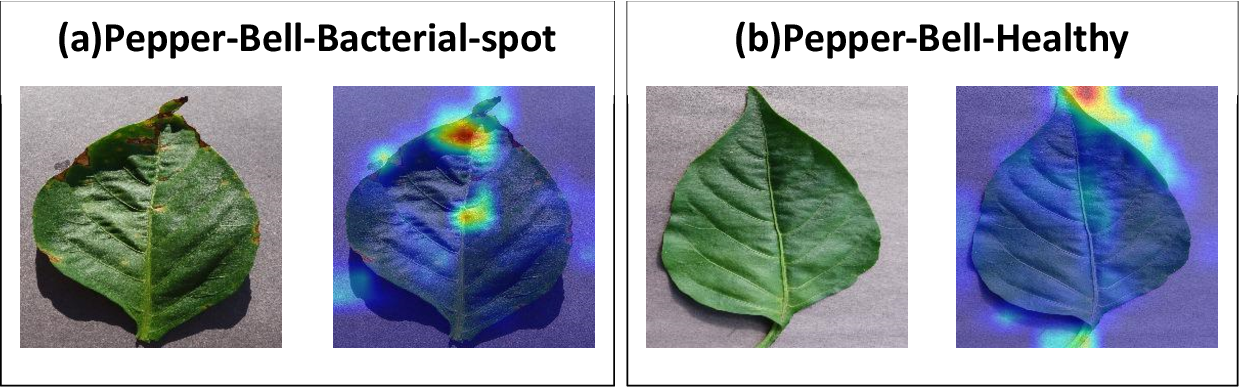}
  \caption{Input images with the obtained features map of all pepper classes }
\end{figure}

\subsubsection{Potato}
The potato category has three classes: one healthy class and two disease classes, Potato-Early-blight and Potato-Late-blight. Heatmaps of all potato classes are presented in Figure 18, which shows the visualization of the proposed MV-CNN model.

\begin{figure}
  \centering
  \includegraphics[width=7cm]{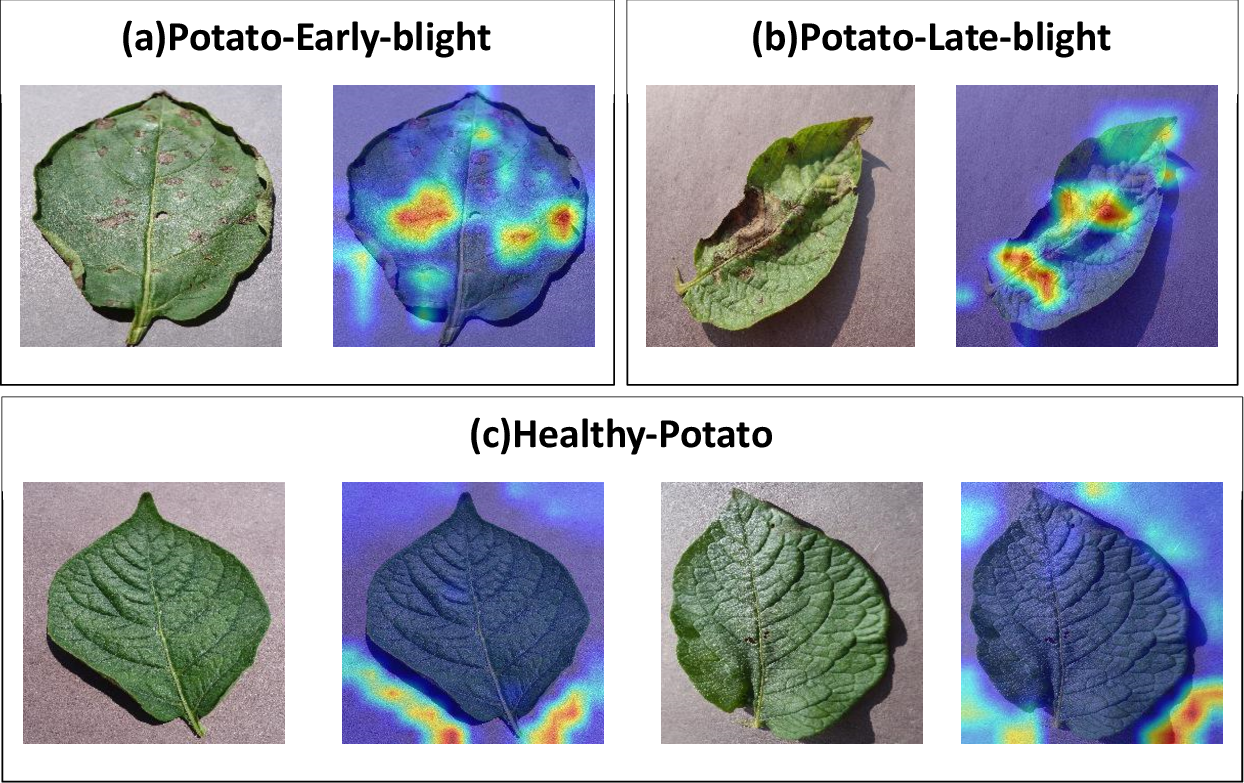}
  \caption{Input images with their calculated feature maps of disease classes of potato, including the healthy class}
\end{figure}

\subsubsection{Raspberry}

This category contains only one class: Raspberry-healthy. To analyze the proposed model's performance, see the feature maps in Figure 19. There is no affected part on the leaf, according to Figure 19.

\begin{figure} 
  \centering
  \includegraphics[width=7cm]{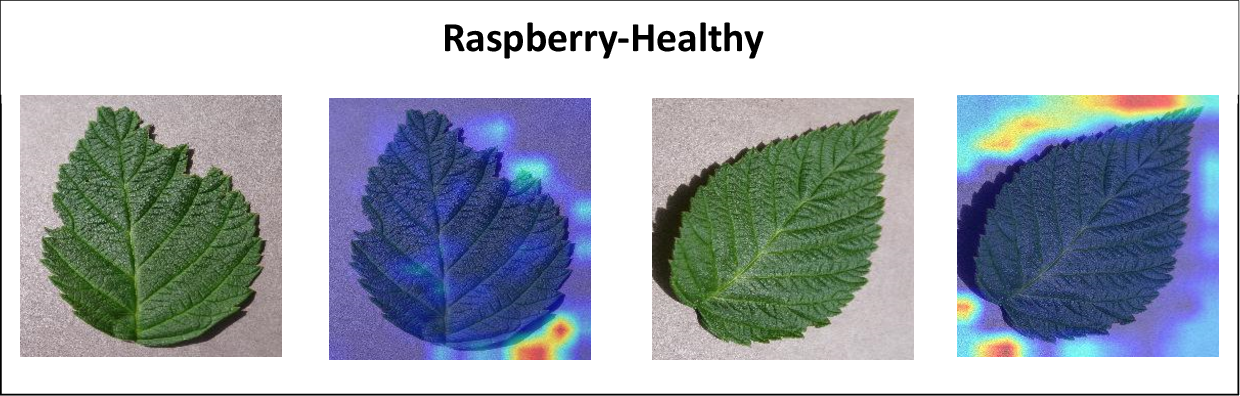}
  \caption{Input images with their calculated feature maps, raspberry healthy class}
\end{figure}

\subsubsection{Soybean}

This is also a healthy class and is categorized as Soybean-healthy. The feature map for this healthy class is shown in Figure 20 to demonstrate the proposed model's performance. It proves that there is no disease in this healthy class.

\begin{figure}
  \centering
  \includegraphics[width=7cm]{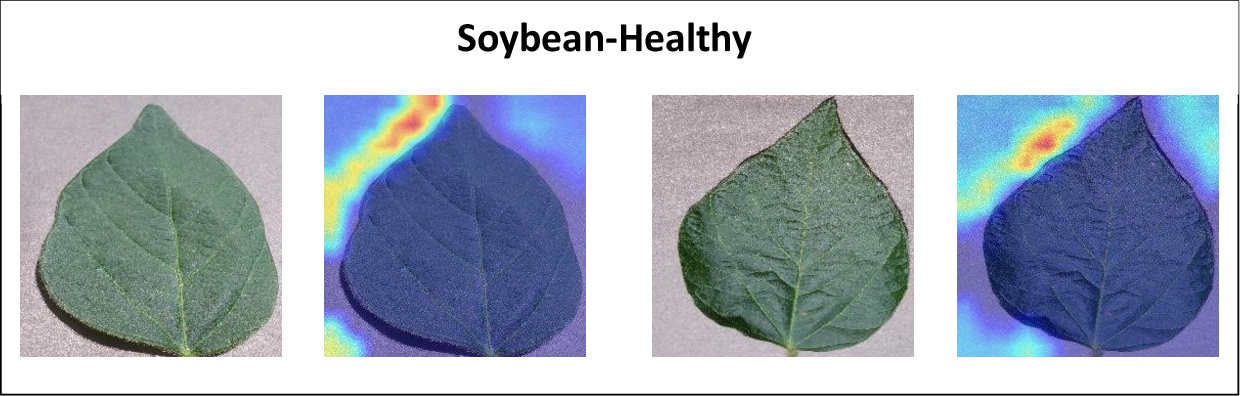}
  \caption{Input images with their calculated feature maps of the healthy class of soybean}
\end{figure}

\subsubsection{Squash}

Squash-Powdery-mildew is a disease of the squash plant and falls into the category of powdery mildew. A more detailed view of this disease with the obtained feature map is presented in Figure 21 to show the proposed MV-CNN model's performance. How the proposed model identifies the disease on the Squash-Powdery-mildew leaf.

\begin{figure}
  \centering
  \includegraphics[width=7cm]{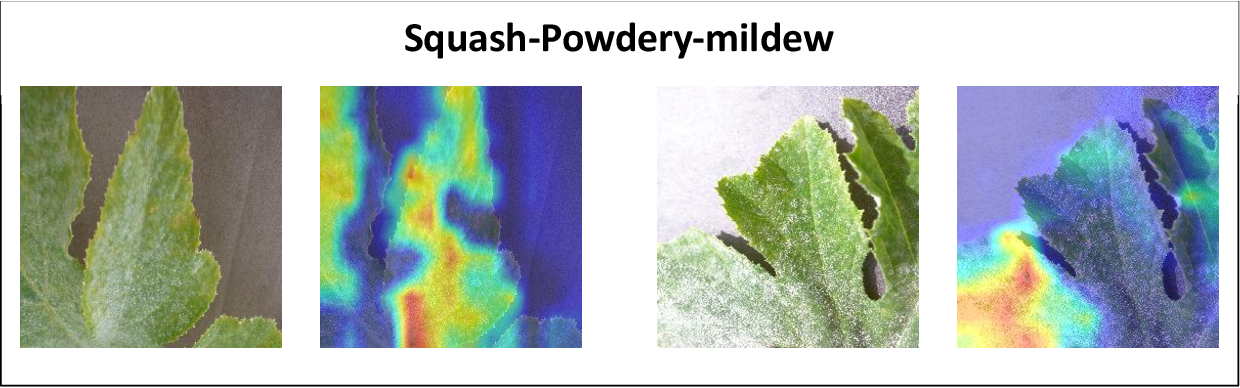}
  \caption{Input images with their calculated feature maps of the squash class}
\end{figure}

\subsubsection{Strawberry}
This is the second-to-last category, with only two classes: Strawberry-Leaf-scorch and Strawberry-healthy. Healthy and affected leaf samples of this category, along with their generated feature maps on the proposed model, are presented in Figure 22. According to the presented feature maps, there is no indication of disease in the healthy class, whereas Strawberry-Leaf-scorch shows some effect.

\begin{figure}
  \centering
  \includegraphics[width=7cm]{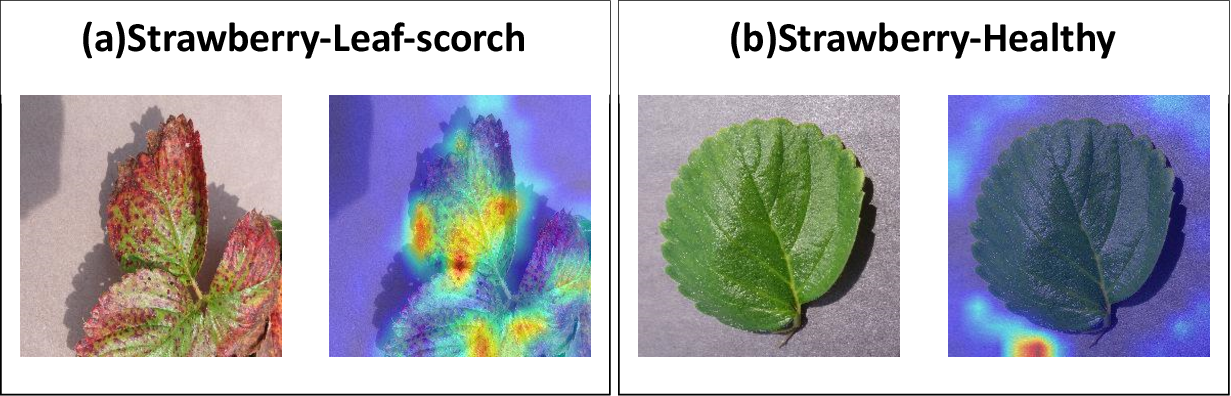}
  \caption{Input images with their calculated feature maps of all classes of strawberry}
\end{figure}

\subsubsection{Tomato}
Tomato is the last and richest category, with the most classes. This category includes 10 classes: Bacterial-spot, Early-blight, Late-blight, Leaf-Mould, Septoria-leaf-spot, and Spider-mites. Two-spotted spider mite is shown with the affected leaf and feature maps in Figure 23. The affected part of each disease is highlighted in this figure.
\begin{figure}
  \centering
  \includegraphics[width=7cm]{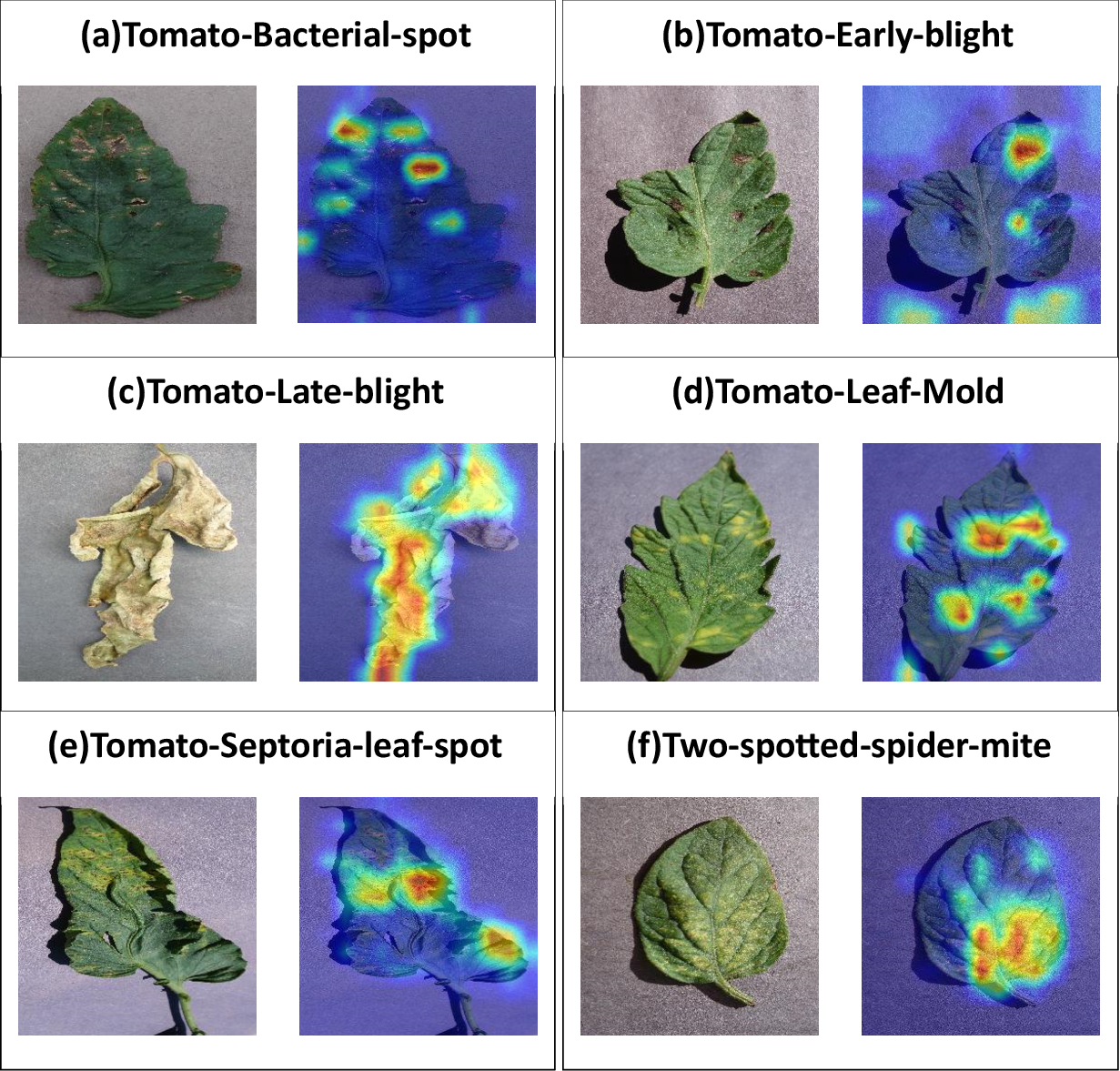}
  \caption{Input images with their calculated feature maps of the first six disease classes of tomato }
\end{figure}
While Target-Spot, Mosaic-virus and Yellow-Leaf-Curl-Virus are in the Tomato-healthy class. These four diseases are presented in Figure 24 to provide a more detailed analysis of tomato disease and their obtained feature maps. The proposed model efficiently identifies the disease on the leaf while the healthy class is out of disease.   
\begin{figure}
  \centering
  \includegraphics[width=7cm]{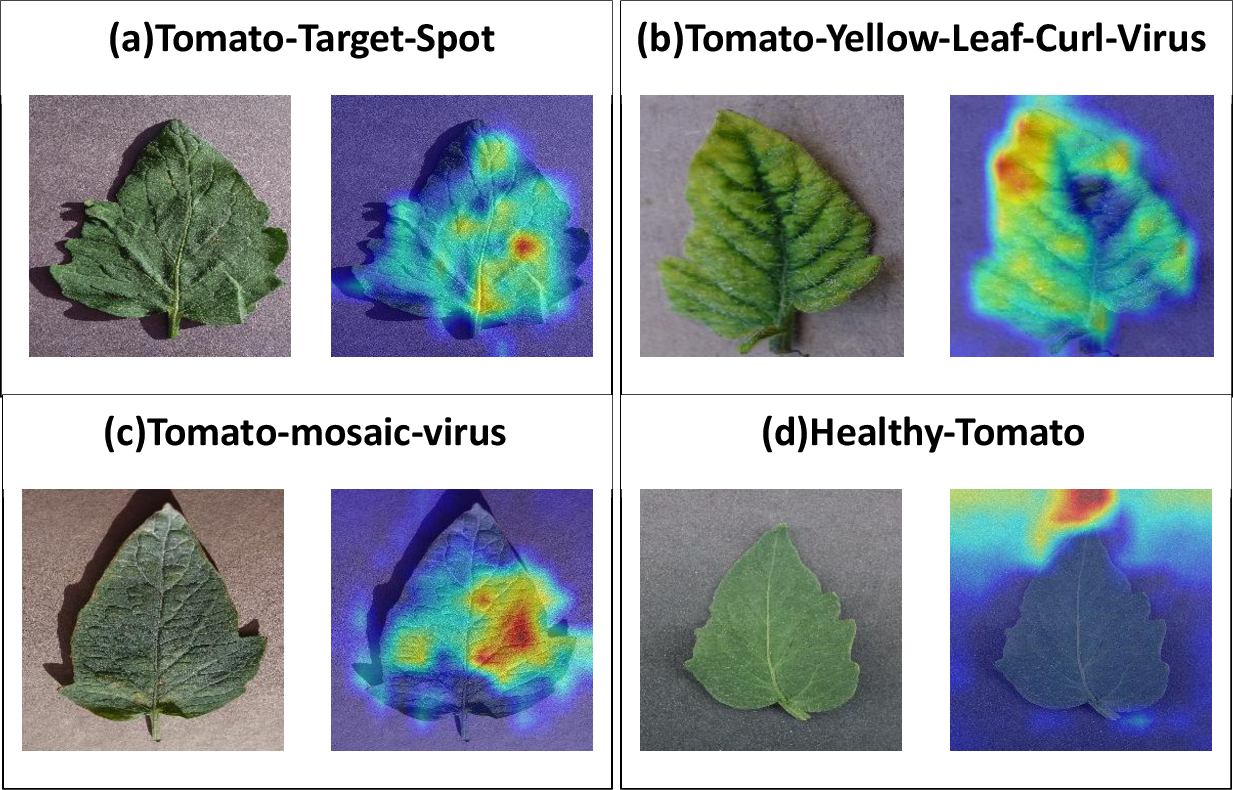}
  \caption{Input images with their calculated feature maps of the last four  disease classes of tomato, including the healthy class}
\end{figure}
These heatmaps show the performance of the proposed MV-CNN model. The proposed model efficiently and accurately identifies the disease on the plant leaf and highlights the affected area. Furthermore, the proposed model also describes that there is no affected region on the healthy leaf. The proposed model highlights the region outside the leaf in healthy classes and gives 100\% class accuracy on all healthy classes. It shows that the proposed model accurately identifies the affected leaves with diseases and healthy leaves.

\subsection{Conclusion}
This article describes the proposed Multi-view CNN model, which uses gradient information as an additional channel with a simple RGB image. Experimentation is based on three types of gradient channels: horizontal, vertical, and gradient magnitude. These additional channels help the network during training and reduce its computational complexity without requiring deeper architectures. The proposed MV-CNN model, with four channels and seven layers, achieves 2.9\% higher accuracy than the baseline nine-layer architecture with three channels on RGB images. In addition, the proposed MV-CNN model achieves results comparable to SOTA classification methods, including Inception-V4, VGGNet, ResNet-50, ResNet-101, DenseNet, and AlexNet. These SOTA networks use RGB images as input, while the proposed MV-CNN model uses gradient magnitude as an additional channel with a simple RGB image. Heatmaps for each class are calculated to show the proposed model's performance, indicating that it can accurately identify the affected part of the leaf. In the case of a healthy class, there is no effect on the leaf. In addition, AUC-ROC curves for each class are calculated for a simple CNN model with only three input channels, the proposed MV-CNN model with four input channels, and ResNet50 with three input channels to demonstrate the performance of the proposed model. AUC-ROC curves and class accuracies indicate that the proposed model performs better overall than a simple CNN. Compared with Resnet50, the proposed model also achieves the same results across all classes and better results in some, such as cherry healthy. The proposed model achieves results comparable to the ResNet50 model with $50$ layers and $23.6$ million parameters, yet uses only $0.726$ million parameters and a simpler architecture. The results and discussions show that an additional gradient channel with an RGB image aids the network in the training and testing phase and improves the network's performance without making the network deep, and gives comparable results with SOTA models.

